%% file: main.tex
\documentclass[final]{cvpr}

\usepackage{times}
\usepackage{epsfig}
\usepackage{graphicx}
\usepackage{amsmath, amsthm, mathtools}
\usepackage{amssymb}
\usepackage{float}
\newtheorem{theorem}{Theorem}[section]

\newtheorem{definition}[theorem]{Definition}

\usepackage[pagebackref=true,breaklinks=true,colorlinks,bookmarks=false]{hyperref}

\def\cvprPaperID{****} 
\def\confYear{CVPR 2021}

\begin{document}

\title{Adversarial Robustness Across Representation Spaces}

\author{Pranjal Awasthi\thanks{Equal contribution and corresponding authors.} \and George Yu$^{*}$ \and Chun-Sung Ferng \and Andrew Tomkins \and Da-Cheng Juan\\
Google Research\\
{\tt\small [pranjalawasthi, georgeyu, csferng, tomkins, dacheng]@google.com}

}

\maketitle

\begin{abstract}
Adversarial robustness corresponds to the susceptibility of deep neural networks to imperceptible perturbations made at test time. In the context of image tasks, many algorithms have been proposed to make neural networks robust to adversarial perturbations made to the input pixels. These perturbations are typically measured in an $\ell_p$ norm. However, robustness often holds only for the specific attack used for training. 
In this work we extend the above setting to consider the problem of training of deep neural networks that can be made simultaneously robust to perturbations applied in multiple natural representations spaces. For the case of image data, examples include the standard pixel representation as well as the representation in the discrete cosine transform~(DCT) basis. We design a theoretically sound algorithm with formal guarantees for the above problem. Furthermore, our guarantees also hold when the goal is to require robustness with respect to multiple $\ell_p$ norm based attacks. We then derive an efficient practical implementation and demonstrate the effectiveness of our approach on standard datasets for image classification.

\end{abstract}

\section{Introduction}
\label{sec:intro}
In recent years deep learning has enjoyed tremendous success in solving a variety of machine learning tasks, even achieving or surpassing human level performance in certain cases~\cite{krizhevsky2009learning, krizhevsky2017imagenet}. At the same time important vulnerabilities in these systems have also been discovered. One such example is their susceptibility to imperceptible perturbations made to the input at test time~\cite{szegedy2013intriguing}. This has led to the new paradigm of {\em adversarial machine learning}, i.e., making deep neural networks robust to test time perturbations. There has been a flurry of recent works in this area with several proposed defenses~\cite{madry2017towards, zhang2019theoretically, cohen2019certified, lecuyer2019certified} and methods to attack and evaluate these defenses~\cite{carlini2017towards, athalye2018obfuscated, tramer2020adaptive}. When studying the design of networks robust to adversarial attacks several aspects need to be considered such as a) what perturbations can the adversary apply to the input, and b) what information does the adversary have about the neural network? One  widely-studied setting in the current literature is {\em white box} attacks under $\ell_p$ norm perturbations~\cite{goodfellow2014explaining, madry2017towards}. Here the adversary has complete knowledge of the neural network and its parameters, and given an input $x$ it can perturb it to $x'$ such that $\|x-x'\|_p \leq \epsilon$ for some $p \geq 1$ specified apriori. In the context of image data this corresponds to applying perturbations to the input pixels. Current approaches for defending against such attacks are based on studying variants of the following robust objective:
\begin{align}
    \label{eq:robust-obj}
    \min_{\theta} \mathbb{E}_{(x,y) \sim D} \big[ \max_{x': \|x-x'\|_p \leq \epsilon} L(f_\theta(x'), y) \big].
\end{align}
Here $(x,y)$ is an example and label pair drawn from the data distribution, $f$ is a neural network parameterized by weights $\theta$ and $L$ is a standard loss function such as the cross entropy loss. As an example the popular projected gradient descent~(PGD) method~\cite{madry2017towards} proposes to optimize the above objective by alternately maximizing the inner objective via gradient ascent and then performing the outer minimization via gradient descent. The recent work of~\cite{salman2019provably} combines the above objective with Gaussian smoothing to achieve {\em certified robustness} guarantees, and another popular method namely the TRADES algorithm~\cite{zhang2019theoretically} adds a regularization term requiring the predictions of the network at $x$ and $x'$ to be close to each other. 

In this work we aim to address two main limitations of current approaches to adversarial machine learning. The first concerns the choice of the representation in which the adversary applies the perturbations. Using images as an example, current approaches model the adversary as making small magnitude changes in the pixel representation of the image. However, given that the adversary has full access to the input $x$, apriori there is no reason to restrict the perturbations to only the pixel representations. Real data such as images have many other natural representations, such as the {\em discrete cosine transform}~(DCT) basis for images. One could envision an adversary making changes to the input image in the DCT basis that are still imperceptible but don't satisfy the small $\ell_p$ norm property in the pixel basis. Empirical attacks based on this have been shown to be successful in recent works~\cite{awasthi2020adversarial}. Hence it is important to consider adversarial robustness in other representations for a model to be truly robust. Secondly, current approaches fix a representation and the perturbation model, and design an algorithm to achieve robustness for that specific setting. In general such networks do not turn out to be robust to other types of attacks. For example a network trained to be robust to $\ell_\infty$ norm perturbations in the pixel representation may not be robust to $\ell_1$ norm perturbations. 

Ideally, one would like to train networks that can be simultaneously robust to multiple attack models in multiple representation spaces. At the same time it is desirable to have a scalable solution with training cost not that much more than standard adversarial training in a fixed attack model. This is precisely the problem that we solve in this work. Our main contributions are listed below.
\begin{itemize}
    \item We propose and motivate the problem of studying robustness to adversarial perturbations in multiple representation spaces and under multiple attack models.
    \item We propose a min-max formulation of the above scenario and use ideas from the theory of online learning, in particular the multiplicative weights update method~\cite{Kalethesis} to design an algorithm for our formulation and provide theoretical guarantees to justify our approach.
    \item We extend our theoretically principled algorithm to design a practical implementation that can scale to  multiple representation spaces and multiple attack models with training cost not significantly more than that of standard adversarial training for a fixed attack model and representation space. We demonstrate the effectiveness of our algorithm for image classification tasks on the MNIST~\cite{lecun1998gradient} and the CIFAR-10~\cite{krizhevsky2009learning} datasets.
\end{itemize}


\section{Related Work}
\label{sec:related}
There is a vast amount of literature on defenses and attacks for adversarial robustness. See~\cite{tramer2020adaptive} for a survey. Here we discuss the works most relevant to the results of the paper. As mentioned in the introduction most existing defenses for adversarial robustness design customized solution for a fixed attack model~($\ell_p$ norm) and representation space~(pixel basis). These methods are aimed at approximately solving the robust optimization objective in \eqref{eq:robust-obj}. The FGSM method~\cite{goodfellow2014explaining} solves the inner maximization problem via one step of a gradient ascent whereas the PGD method~\cite{madry2017towards} performs multiple iterations of gradient ascent to better optimize the inner objective. Typically this scales the cost of training linearly with the number of iterations used in the inner maximization. There have been recent works aimed at achieving the same performance as the PGD method but  with faster training time~\cite{shafahi2019adversarial, wong2020fast}.

The above approaches provide robustness to first order attacks that are of the same type that are used in training. There has also been a lot of recent work on {\em provably} certifying the robustness of neural networks via approaches such as interval bound propagation~\cite{gowal2018effectiveness}, semi-definite programming~\cite{raghunathan2018semidefinite}, and randomized smoothing~\cite{cohen2019certified, lecuyer2019certified}. 

Relatively little work exists on studying robustness to multiple types of attacks simultaneously and in multiple representation spaces. The recent work of~\cite{tramer2019adversarial} studies training classifiers that are simultaneously robust to perturbations to the input pixels of different $\ell_p$ norms. However they do not consider multiple representation spaces. Furthermore, their approach does not come with theoretical guarantees and scales linearly with the number of perturbations considered. In contrast our algorithm comes with theoretical guarantees and has a training cost that is not much more than that of adversarial training for a fixed attack. The recent work of~\cite{awasthi2020adversarial} motivates the problem of studying certified robustness in other representations such as the DCT basis. However they do not consider training classifiers that are simultaneously robust to multiple attack models.


\section{Adversarial Robustness in Multiple Representations}
\label{sec:multi-rep}
In this section we motivate the need for studying adversarial robustness in representations other than the one that is input to the network. Real world data can be represented in many natural representations, each with their own appealing properties. For instance, in the context of images, the DCT basis is a popular choice and it is well known that signals when represented in this basis are sparse. This has been exploited in recent works~\cite{awasthi2020adversarial} to achieve better robustness to $\ell_\infty$ perturbations in this representations.
\begin{figure*}[htbp]
    \centering
    \includegraphics[width=0.5\textwidth]{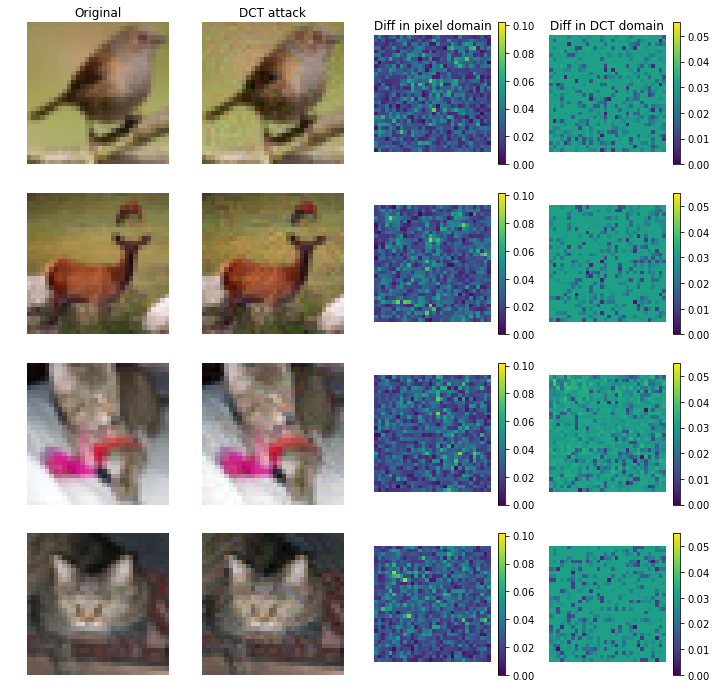}
    \caption{The figure shows examples of images from the CIFAR-10 dataset with their adversarially perturbed counterparts computed by launching a PGD based attack in the DCT basis. The perturbed images, although imperceptible, are far from the original images in the pixel basis in $\ell_\infty$ norm.}
    \label{fig:perturbations}
\end{figure*}

In the context of adversarial learning a fundamental question to ask is: {\em what constitutes an imperceptible perturbation?} Is it enough for an adversarially perturbed example to have a small $\ell_p$ norm in the pixel representation? As one can imagine, this is not a sufficient condition for imperceptibility. Many works have notice that images and their adversarial perturbations made in the pixel basis have distinct spectral signatures when views in other bases such as the discrete cosine transform~(DCT). This has led to the proposal of many learning systems for detecting pixel based adversarial attacks using properties of images in other representations~\cite{akhtar2018defense, yin2019fourier, dziugaite2016study}. In particular, the work of~\cite{metzen2017detecting} shows that one can achieve high accuracy in detecting adversarial perturbations made in the pixel representation by training a binary classifier to separate real and perturbed images. Hence an adversary has to naturally think about attacking the model in multiple representations simultaneously in order to fool such systems.

Additionally, working in multiple representation spaces can help an adversary craft stronger attacks. As an example, the recent work of~\cite{awasthi2020adversarial} provides examples where one can generate imperceptible examples by perturbing the image in the DCT basis and at the same time the perturbed examples are far way from the original image in the original pixel basis. Such an attack can fool classifiers that are only trained for defending against small norm $\ell_p$ attacks in the pixel representation. We further illustrate this in Figure~\ref{fig:perturbations}. The figure shows examples of images from the CIFAR-10 dataset and corresponding adversarial perturbations computed by launching a PGD based adversarial attack in the DCT basis. For the case of CIFAR-10 it is generally accepted that $\ell_\infty$ perturbations in the pixel basis upto a magnitude of $\epsilon = 0.03$ constitute imperceptible perturbations. However, the adversarial images obtained in the Figure via working in the DCT basis, while being imperceptible, have much higher $\ell_\infty$ distance from the true images in the pixel representation.

From the above discussion we conclude that it is an important problem to design classifiers that are simultaneously robust against adversarial attacks in multiple representation spaces. Unfortunately, simply performing standard adversarial training in a fixed space is not enough in order to achieve this goal. As an example in Table~\ref{table:tradeoff} we show the performance of two neural networks, one trained adversarially in the pixel representation and the other in the DCT representation. As can be seen the trained networks have very poor robustness against the attacks that were not considered during training.


\begin{table}[htbp]
\small
    \centering
    \begin{tabular}{|p{1.6cm}|p{1.6cm}|p{1.6cm}|p{1.6cm}|}
    \hline
         &  Test w/pixel $\ell_\infty$ & Test w/DCT $\ell_\infty$ & Nat. Acc.\\
         \hline
         Train w/pixel $\ell_\infty$ & $44.02$ $\pm 1.02$ & $27.30$ $\pm 1.44$  & $80.24$ $\pm 0.40$\\
         \hline
         Train w/DCT $\ell_\infty$ & $11.80$ $\pm 0.68$ & $51.92$ $\pm 0.43$  & $74.92$ $\pm 0.61$\\
         \hline
    \end{tabular}
    \caption{The rows of the table correspond to two classifiers that have been adversarially trained via the PGD method for $\ell_\infty$ robustness either in the pixel basis or the DCT basis. The first two columns show the adversarial accuracies achieved by the classifiers against $\ell_\infty$ attacks in the pixel and the DCT basis. The last column displays the natural accuracy. As can be seen no classifier is simultaneously robust to both types of attacks.}
    \label{table:tradeoff}
\end{table}
As a result of the above observations what is needed is a general algorithmic approach for such scenarios. We next formulate and present such an approach.


\section{Formulation and Algorithms}
\label{sec:alg}
We next formulate the above scenario and design a near optimal algorithm for simultaneously achieving robustness across multiple representation spaces. We fix a canonical representation~(say the pixel basis) and denote $x \in \mathbb{R}^d$ as examples and $y$ being the label. We assume that the example and label pairs $(x,y)$ are drawn from an unknown joint distribution $D$. We then consider a given set of $k$ representation spaces with corresponding maps being given by $\mathcal{R}_1, \mathcal{R}_2, \dots, \mathcal{R}_k$. Hence, given an example $x \in \mathbb{R}^d$ its representation in space $i$ is given by $\mathcal{R}_i(x) \in \mathbb{R}^{d_i}$. It would be instructive to think of $\mathcal{R}_i$ as the DCT basis although in general these maps could be non-linear. The only assumption we require is that the maps be surjective, i.e. $\mathcal{R}^{-1}_i$ exists. We will overload notation and denote by $\mathcal{R}_i$ both the $i$th representation and the map corresponding to it. For each $\mathcal{R}_i$ and any $x \in \mathbb{R}^d$, we denote by $B_i(x)$ the set of allowed perturbations to $x$ in the representation space $\mathcal{R}_i$. For example if we are modeling an $\ell_\infty$ attack of radius $\epsilon$ in the representation $\mathcal{R}_i$ then we have $B_i(x) = \{z \in \mathbb{R}^{d_i}: \|\mathcal{R}_i(x) - z\|_\infty \leq \epsilon\}$. In this work we will be able to deal with very general perturbation sets. Given a fixed representation space $\mathcal{R}_i$, the problem of learning a robust classifier specific to $\mathcal{R}_i$ can be written as that of minimizing:
\begin{align}
    \label{eq:robust-obj-i}
    \min_{\theta} L_i(\theta) =
    \mathbb{E}_{(x,y) \sim D} \big[ \max_{z \in B_i(x)} L(f_\theta(\mathcal{R}^{-1}_i(z)), y) \big].    
\end{align}

Then given $k$ representation spaces $\mathcal{R}_1, \dots , \mathcal{R}_k$ our goal is to solve the following:
\begin{align}
    \label{eq:robust-obj-simultneous}
    \min_{\theta} \max_i L_i(\theta).
\end{align}

The above min-max formulation lends itself naturally to techniques from online learning. In particular, consider a two player game with the row player as the one that chooses the network parameter $\theta$ and the column player as the one that chooses the loss functions $L_i$ with the payoff for the column player being $L_i(\theta)$. From the minimax theorem~\cite{Kalethesis} for two player games, we know that if for every distribution over the $k$ columns there exists a good solution $\theta$, then there exists a distribution over solutions that is simultaneously good for all the columns, i.e., the $k$ loss functions. This immediately provides a way to solve the min-max formulation via solving a simple {\em cost sensitive} adversarial optimization problem. Such techniques have been widely used in the literature to solve a variety of constrained problems in machine learning~\cite{agarwal2018reductions, cotter2019two}. Here we demonstrate their applicability for adversarial robustness. There has also been recent work on algorithms for solving optimization of the form $\min_{\theta} \max_{\lambda \in \Lambda} \sum_i \lambda_i L_i(\theta)$~\cite{cortes2020agnostic} for $\Lambda$ being a convex set and functions $L_i$ being convex in $\theta$. Our loss functions are non-convex in $\theta$ and hence we need access to a cost sensitive optimization oracle to provide overall guarantees for our formulation. 
We next define the {\em adversarial cost sensitive} optimization problem.
\begin{definition}
Given weights $w_1, w_2, \dots, w_k$ with $w_i \geq 0$ and non-negative losses $L_1, L_2, \dots, L_k$ the adversarial cost sensitive optimization corresponds to finding an approximately optimal solution $\hat{\theta}$ such that
\begin{align}
    \label{eq:robust-cost-sensitive}
    \sum_{i=1}^k w_i L_i(\hat{\theta}) \leq  \min_\theta \sum_{i=1}^k w_i L_i({\theta}) + \delta.
\end{align}
Here $\delta$ quantifies the additive error in approximating the cost sensitive objective.
\end{definition}

\begin{figure*}[t]
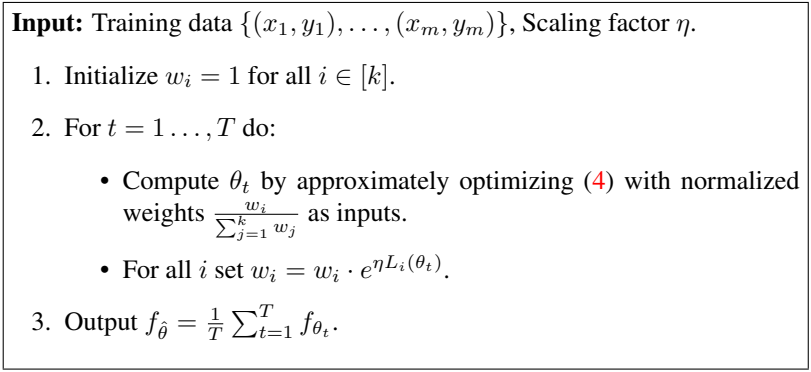

\begin{center}
\fbox{\parbox{0.6\textwidth}{
{\bf Input:} Training data $\{(x_1, y_1), \dots, (x_m, y_m)\}$, Scaling factor $\eta$. 
\begin{enumerate}   
\item Initialize $w_i=1$ for all $i \in [k]$.

\item For $t=1 \dots, T$ do:
\begin{itemize}
    \item Compute $\theta_t$ by approximately optimizing \eqref{eq:robust-cost-sensitive} with normalized weights $\frac{w_i}{\sum_{j=1}^k w_j}$ as inputs. 
    \item For all $i$ set $w_i = w_i \cdot e^{\eta {L}_{i}(\theta_t)}$.
\end{itemize}
\item Output $f_{\hat{\theta}} = \frac{1}{T} \sum_{t=1}^T f_{\theta_t}$.
\end{enumerate}}}
\end{center}
\caption{An algorithm achieving robustness simultaneously across representation spaces.}
\label{ALG:multi-robust} 
\end{figure*}

We will show how to convert an algorithm for solving the adversarial cost sensitive optimization problem above to provably optimize \eqref{eq:robust-obj-simultneous}. The algorithm is based on the popular multiplicative weights update method~\cite{Kalethesis} and is described in Figure~\ref{ALG:multi-robust}.
For the proposed algorithm we show the following guarantee
\begin{theorem}
\label{thm:main}
For a given set of non-negative losses bounded in $[0,R]$, if the adversarial cost sensitive optimization in \eqref{eq:robust-cost-sensitive} can be solved to additive error $\delta$ for any setting of non-negative weights then the algorithm in Figure~\ref{ALG:multi-robust} when run with $\eta = O(\epsilon/R)$ and $T = O(\frac{R^2 \log k}{\epsilon^2})$ outputs a uniform distribution $P$ over solutions $\theta_1, \theta_2, \dots, \theta_T$ such that
\begin{align}
    \label{eq:guarantee}
    \max_i \mathbb{E}_{\theta \sim P} L_i(\theta) \leq \min_\theta \max_i L_i(\theta) + \epsilon + \delta.
\end{align}
Furthermore if the loss function $L$ in \eqref{eq:robust-obj-i} is convex in its first argument, such as the cross-entropy loss, squared loss and hinge loss to name a few, then the average hypothesis $f_{\hat{\theta}}$ satisfies
\begin{align}
    \label{eq:guarantee-avg}
    \max_i L_i(f_{\hat{\theta}}) \leq \min_\theta \max_i L_i(\theta) + \epsilon + \delta.
\end{align}
Here $L_i(f_{\hat{\theta}})$ refers to the loss incurred by the ensembled hypothesis as output by the algorithm in Figure~\ref{ALG:multi-robust}.
\end{theorem}

The proof can be found in Appendix~\ref{app:proof} in the supplementary material.

\section{A Practical Implementation}
\label{sec:practical}
While the algorithm in Figure~\ref{ALG:multi-robust} and the associated guarantee in Theorem~\ref{thm:main} provide a principled way to approach the optimization in \eqref{eq:robust-obj-simultneous}, we need to make a number of modifications to the core algorithm in order to obtain a practical and scalable implementation. In particular, we do not want the cost of training the robust classifiers to scale linearly with $k$ the number of representation spaces. We first discuss solving the adversarial cost sensitive optimization in \eqref{eq:robust-cost-sensitive}. In practice, each $L_i(\theta)$ itself represents a hard optimization problem~(of the form \eqref{eq:robust-obj-i}). Luckily, there exists first order algorithms such as the PGD method~\cite{madry2017towards} to optimize each $L_i$ separately that work well in practice. We now show how to combine them to solve \eqref{eq:robust-cost-sensitive}. We follow the methodology of stochastic optimization and proceed in epochs. In each epoch, we sample a mini batch of $B$ data points, sample a loss $L_i$ with probability proportional to its current $w_i$ and then run the corresponding PGD based algorithm for optimizing $L_i$ on the current mini batch. After a few epochs of optimization we update the weights $w_i$ of the losses as described in the algorithm in Figure~\ref{ALG:multi-robust}. In order to evaluate the losses for the weight update we use a separate validation set. This significantly reduces the variance in our estimates.

Next we consider approximating the output $f_{\hat{\theta}}$. Notice that the guarantee of Theorem~\ref{thm:main} applies to an ensemble of $T$ neural networks provided by parameters $\theta_1, \theta_2, \dots, \theta_T$. Maintaining this ensemble requires a high storage cost and makes the final output model impractically big. We first notice that if the losses $L_i$ were convex, then the guarantee of Theorem~\ref{thm:main} will also hold for the average parameter, i.e., $\hat{\theta} = \frac{1}{T} \sum_t \theta_t$. To get a practical implementation we make a {\em near convexity} assumption on the losses and simply take the average of the model weights. Furthermore, in our experiments we observe that taking the average of the last few model parameters performs better than the uniform average of all the model weights. Fixing these choices leads to a scalable variant as shown in Figure~\ref{ALG:multi-robust-scalable}.
\begin{figure*}[t]
\begin{center}
\fbox{\parbox{1\textwidth}{
{\bf Input:} Training data $\{(x_1, y_1), \dots, (x_m, y_m)\}$, Validation data $\{(x_{m+1}, y_{m+1}), \dots, (x_{m+s}, y_{m+s})\}$, mini batch size $B$, time steps $T$, update frequency $r$, window size $h$, Scaling factor $\eta$. 
\begin{enumerate}   
\item Initialize $w_i=1$ for all $i \in [k]$.

\item For $t=1 \dots, T$ do:
\begin{itemize}
    \item Repeat for $r$ epochs:
    \begin{itemize}
    \item Get the next mini batch of size $B$. Sample loss $L_i$ with probability $p_i = \frac{w_i}{\sum_{j=1}^k w_j}$.
    \item Run the PGD based algorithm to optimize $L_i$ on the mini batch.
\end{itemize}
\item For all $i$ set $w_i = w_i \cdot e^{\eta {L_i^{\text{val}}}(\theta_t)}$. Here $L^{\text{val}}$ is the loss evaluated on the validation set.
\end{itemize}
\item Output $\hat{\theta} = \frac{1}{h} \sum_{t=T-h+1}^T \theta_t$.
\end{enumerate}}}
\end{center}
\caption{A scalable variant of the algorithm in Figure~\ref{ALG:multi-robust}.}
\label{ALG:multi-robust-scalable} 
\end{figure*}

\section{Experimental Evaluation}
\label{sec:experiments}
\begin{figure}[htbp]
    \centering
    \includegraphics[width=0.15\textwidth]{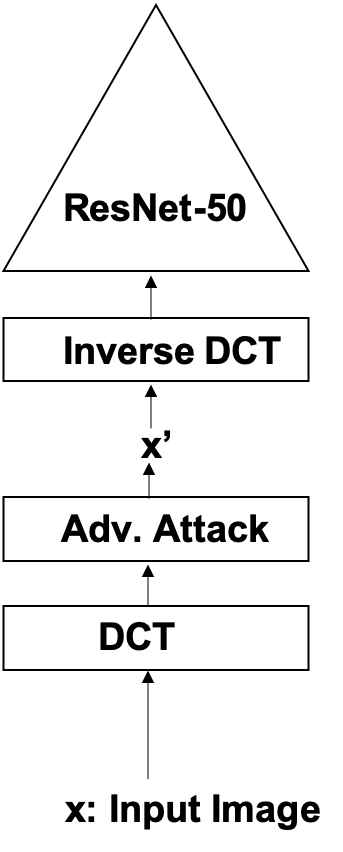}
    \caption{The modified network architecture for computing an adversarial perturbation in the DCT basis.}
    \label{fig:dct-network}
\end{figure}

We next demonstrate the effectiveness of our approach on the task of learning a classifier for image classification that is simultaneously robust to adversarial attacks in multiple representations spaces. 
\paragraph{Datasets.} We perform the experimental evaluation on two public datasets namely the MNIST dataset~\cite{lecun1998gradient} and the CIFAR-10 dataset~\cite{krizhevsky2009learning}. The MNIST dataset consists of $60,000$ training images with each being a $28 \times 28 \times 1$ tensor. The CIFAR-10 dataset consists of $50,000$ training images each of dimensionality $32 \times 32 \times 3$. 
Both the datasets consist of $10,000$ test images and correspond to a multi class classification problem with $10$ class labels. In each case we reserve $10\%$ of the training data to be used as the validation set in the Algorithm from Figure~\ref{ALG:multi-robust-scalable}. This validation set will be used to evaluate the loss $L^{\text{val}}$ in the algorithms.
\vspace{-5mm}
\paragraph{Representation Spaces and Attack Models.} To demonstrate the scalability of our approach we consider two different representation spaces namely the pixel basis and the DCT basis. In each representation space we consider three types of $\ell_p$ norm based attacks for $p=1,2,\infty$. Hence, in total we have $6$ loss functions $L_i$ to optimize as in \eqref{eq:robust-obj-simultneous}. All our experiments are conducted on a ResNet-50 deep neural network that is a popular architecture for training on image classification tasks~\cite{he2016deep}. To compute adversarial examples in the pixel basis for norm bounded $\ell_\infty$ and $\ell_2$ perturbations we use the standard PGD based attack as proposed in~\cite{madry2017towards}. For computing a norm bounded $\ell_1$ perturbation we use the sparse ascent algorithm namely the SLIDE method as proposed in~\cite{tramer2019adversarial}. To compute an adversarial attack in the DCT basis  we append the ResNet-50 architecture with a linear DCT transformation follows by an inverse DCT transformation as shown in Figure~\ref{fig:dct-network}. Notice that both the DCT and the Inverse DCT are fixed linear layers and in the absence of any perturbations the output of the network in Figure~\ref{fig:dct-network} is exactly the same as the original ResNet-50 network. 

Using the modified architecture we first compute the DCT representation of the image and then launch an adversarial perturbation in the DCT basis using either the PGD method~(for $\ell_2, \ell_\infty$ attacks) or the SLIDE method~(for $\ell_1$ attacks). In this way we get the perturbed image after taking the inverse DCT transform of the perturbed example $x'$ as shown in Figure~\ref{fig:perturbations}. After computing the adversarial perturbation and passing it through the inverse DCT transform we clip the pixel values in $[0,1]$ to make the example a valid input for the ResNet-50 network.

\vspace{-5mm}
\paragraph{Baselines.} We next describe the baselines that we use when comparing our proposed approach. The problem of being simultaneously robust to multiple adversarial attacks has been largely ignored in the literature so far. The recent work of~\cite{tramer2019adversarial} studies being robust in pixel basis to different $\ell_p$ norm based attacks. However the proposed method is not scalable to a large number of attacks.

We instead compare our proposed algorithm with following two baseline heuristics. We choose these heuristics due to their simplicity and more importantly due to their scalability. 

\noindent \textbf{Round Robin.} The round robin heuristic sketched in Figure~\ref{ALG:round-robin} follows the same outline as the algorithm in Figure~\ref{ALG:multi-robust-scalable} except that instead of maintaining and updating weights, it simply picks the loss function $L_i$ to apply to a mini batch in a fixed order. It is easy to see that this method scales very well. 
\begin{figure}[htbp]
\begin{center}
\fbox{\parbox{0.4\textwidth}{
{\bf Input:} Training data $\{(x_1, y_1), \dots, (x_m, y_m)\}$.

{\bf Input:} Mini batch size $B$, time steps $T$. 
\begin{enumerate}   
\item Initialize $index=1$.

\item For $t=1 \dots, T$ do:
\begin{itemize}
    \item Get the next mini batch of size $B$. 
    \item Set $i = index$. 
    \item Use PGD to optimize $L_i$ on the mini batch to get $\theta_t$.
    \item $index = (index+1) \text{ mod } k + 1$.
\end{itemize}
\item Output $\hat{\theta} = \theta_T$.
\end{enumerate}}}
\end{center}
\caption{The round robin heuristic.}
\label{ALG:round-robin} 
\end{figure}

\noindent \textbf{Greedy.} The greedy heuristic also follows the same outline as the algorithm in Figure~\ref{ALG:multi-robust-scalable}. However, for each time step it chooses the loss with the worst error~(on the validation set) to apply next. See Figure~\ref{ALG:greedy}.
\begin{figure}[htbp]
\begin{center}
\fbox{\parbox{0.4\textwidth}{
{\bf Input:} Training data $\{(x_1, y_1), \dots, (x_m, y_m)\}$.

{\bf Input:} Validation data: 

$\{(x_{m+1}, y_{m+1}), \dots, (x_{m+s}, y_{m+s})\}$.

{\bf Input:} Mini batch size $B$, time steps $T$, update frequency $r$. 
\begin{enumerate}   
\item For $t=1 \dots, T$ do:
\begin{itemize}
\item Set $i = \arg \max_j L^{\text{val}}_j$.
\item Repeat for $r$ epochs:
\begin{itemize}
    \item Get the next mini batch of size $B$. 
    \item Use PGD to optimize $L_i$ on the mini batch to get $\theta_t$.
\end{itemize}
\end{itemize}
\item Output $\hat{\theta} = \theta_T$.
\end{enumerate}}}
\end{center}
\caption{The greedy heuristic.}
\label{ALG:greedy} 
\end{figure}

\begin{table*}[t]
\small
    \centering
    \begin{tabular}{|c|c|c|c|c|}
    \hline
         &  Greedy & Round Robin & Mult. Weights~($h=1$) & Mult. Weights~($h=3$)\\
         \hline
         Pixel ($\ell_\infty$) & $36.03	\pm 6.69$ & $29.99	\pm 1.49$ & ${35.70	\pm 6.36}$ & $\mathbf{39.13	\pm 1.74}$\\
         \hline
         Pixel ($\ell_2$) & $69.03	\pm 1.58$ & $\mathbf{71.92	\pm 0.36}$ & $69.51	\pm 2.21$ & $\mathbf{71.27	\pm 0.22}$\\
         \hline
         Pixel ($\ell_1$) & $45.84	\pm 2.85$ & $\mathbf{53.45	\pm 0.70}$ & $44.22 \pm	4.79$ & $46.90	\pm 1.46$\\
         \hline
         DCT ($\ell_\infty$) & ${44.72	\pm 8.60}$ & $\mathbf{45.63	\pm 1.75}$ & $39.44	\pm 6.76$ & $42.27	\pm 2.77$\\
         \hline
         DCT ($\ell_2$) & $68.99	\pm 1.58$ & $\mathbf{72.01	\pm 0.28}$ & $69.66 \pm	2.21$ & $71.24	\pm 0.12$\\
         \hline
         DCT ($\ell_1$) & $37.51	\pm 6.78$ & $\mathbf{41.44	\pm 0.82}$ & $39.62	\pm 5.71$ & $\mathbf{42.96	\pm 1.12}$\\
         \hline
         Min. Accuracy & $34.62	\pm 5.60$ & $29.99	\pm 1.49$ & $\mathbf{35.70	\pm 6.36}$ & $\mathbf{39.13	\pm 1.74}$\\
         \hline
         Union Attack  & $31.95	\pm 4.55$ & $29.70	\pm 1.42$ & $\mathbf{33.17	\pm 5.95}$ & $\mathbf{36.18	\pm 1.89}$\\
         \hline
         Nat. Acc. & $78.56	\pm 1.46$ & $\mathbf{81.80	\pm 0.38}$ & $79.64	\pm 0.51$ & $\mathbf{80.66	\pm 0.89}$\\
         \hline
    \end{tabular}
    \caption{Comparison of the adversarial accuracies achieved on the CIFAR-10 dataset by the greedy algorithm, the round robin algorithm and our proposed algorithm in Figure~\ref{ALG:multi-robust-scalable}.}
    \label{table:acc-cifar10}
\end{table*}
\vspace{-5mm}
\paragraph{Hyperparameter Configurations.} Next we discuss the hyperparameters we use when computing the adversarial perturbations for the different $\ell_p$ norm based attacks. When running our proposed algorithm in Figure~\ref{ALG:multi-robust-scalable} and the greedy heuristic, we set $T=40$, $r=5$ for CIFAR-10~($200$ epochs total), and $T=20$, $r=3$ for MNIST~($60$ epochs total). We train the round robin heuristic for the same number of epochs. 

For the case of $\ell_\infty$ and $\ell_2$ attacks, during training we use $10$ steps of gradient ascent to optimize the inner maximization in \eqref{eq:robust-obj-i}. During evaluation we again run the PGD based attack on our model across all the representation spaces and use $40$ steps of the PGD method to solve the inner maximization in \eqref{eq:robust-obj}. For the case of $\ell_1$ attacks we use $20$ iterations of the SLIDE method during training to compute adversarial perturbations and $100$ iterations of the method during evaluation. We experiment with both running the PGD method with $20$ random restarts, and a simpler attack with no restarts. The experiments we report here are for the latter case. The qualitative conclusions of our experiments remain the same when using $20$ random restarts. See the supplementary material for details. 
\begin{table*}[h]
\small
    \centering
    \begin{tabular}{|c|c|c|c|c|c|}
    \hline
         Dataset & Pixel~($\ell_\infty$) & DCT~($\ell_\infty$) & Round Robin & Greedy & Mult. Weights \\
         \hline
         MNIST & $2.25$ & $2.09$ & $3.36$ & $2.67$ & $2.90$\\
         \hline
         CIFAR-10 & $7.66$ & $7.06$ & $10.61$ & $8.70$ & $9.51$\\
         \hline
    \end{tabular}
    \caption{Training time in hours~(wall clock time) for the baselines and our proposed method on the MNIST and the CIFAR-10 datasets. The first two columns represent the training time for optimizing a single loss, i.e., $\ell_\infty$ attack in the pixel and the DCT basis respectively. The next three columns represent the training time of the three methods when optimizing over all the $6$ losses simultaneously. The reported numbers are averaged over $5$ runs.}
    \label{tab:train-time}
\end{table*}

For the MNIST dataset we perturbation magnitudes of $0.4, 1$ and $5$ for $\ell_\infty$, $\ell_2$ and $\ell_1$ norm based attacks respectively. The corresponding magnitudes for the CIFAR-10 dataset are 
$0.06, 0.1$ and $7.84$. We keep the perturbation magnitudes the same across both the pixel and the DCT basis. In our experiments when performing gradient ascent for $\ell$ steps to compute a perturbation, we use a step size of $2.5 \frac{\epsilon}{\ell}$, where $\epsilon$ is the perturbation magnitude. 
\vspace{-5mm}
\paragraph{Metrics.} For each of our trained classifiers we report the adversarial accuracy for each of the $6$ individual attacks launched separately on the trained model. In addition we also report the worst adversarial accuracy among the $6$ attacks on the same model. Notice that this is the metric that our proposed algorithm in Figure~\ref{ALG:multi-robust-scalable} aims to optimize. Finally, we also report the accuracy of our trained models on a {\em union attack}, i.e., for each example we produce all $6$ adversarial perturbations and consider the attack successful if any one of them succeeds in making the prediction of the model incorrect. Finally, notice that our proposed algorithm in Figure~\ref{ALG:multi-robust-scalable} has  a parameter $h$ namely the window size. We report results for $h=1$ and $h=3$. These correspond to either using  the parameters of the last time step or using the model averaged over the last three time steps. 
\begin{table*}[h]
\small
    \centering
    \begin{tabular}{|c|c|c|c|c|}
    \hline
         &  Greedy & Round Robin & Mult. Weights~($h=1$) & Mult. Weights~($h=3$)\\
         \hline
         Pixel ($\ell_\infty$) & $63.12	\pm 19.50$ & $64.40	\pm 25.12$ & $\mathbf{67.73	\pm 14}$ & $\mathbf{66.56	\pm 13.45}$\\
         \hline
         Pixel ($\ell_2$) & $22.51	\pm 17.66$ & $10.23	\pm 8.40$ & $\mathbf{71.23	\pm 2.68}$ & $\mathbf{71.77 \pm	4.25}$\\
         \hline
         Pixel ($\ell_1$) & $64.33	\pm 30.73$ & $43.14	\pm 23.57$ & $\mathbf{73.20	\pm 10.54}$ & $\mathbf{73.66	\pm 9.47}$\\
         \hline
         DCT ($\ell_\infty$) & $61.53	\pm 21.45$ & $59.47	\pm 32.83$ & $\mathbf{60.65	\pm 10.60}$ & $\mathbf{60.32	\pm 11.34}$\\
         \hline
         DCT ($\ell_2$) & $25.80	\pm 18.82$ & $12.28	\pm 12.16$ & $\mathbf{84.58	\pm 2.52}$ & $\mathbf{84.46	\pm 4.18}$\\
         \hline
         DCT ($\ell_1$) & $66.00	\pm 31.83$ & $38.23	\pm 21.47$ & $\mathbf{72.39	\pm 8.43}$ & $\mathbf{73.13	\pm 7.33}$\\
         \hline
         Min. Accuracy & $22.13	\pm 17.17$ & $9.76	\pm 8.69$ & $\mathbf{57.64	\pm 7.83}$ & $\mathbf{57.57	\pm 8.46}$\\
         \hline
         Union Attack  & $12.32	\pm 9.80$ & $3.72	\pm 4.96$ & $\mathbf{35.30	\pm 4.48}$ & $\mathbf{35.87	\pm 6.69}$\\
         \hline
         Nat. Acc. & $77.16	\pm 37.64$ & $60.93	\pm 32.31$ & $\mathbf{91.00	\pm 10.50}$ & $\mathbf{91.43	\pm 9.61}$\\
         \hline
    \end{tabular}
    \caption{Comparison of the adversarial accuracies achieved on the MNIST dataset by the greedy algorithm, the round robin algorithm and our proposed algorithm in Figure~\ref{ALG:multi-robust-scalable}.}
    \label{table:acc-mnist}
\end{table*}
\vspace{-5mm}
\paragraph{Results.} The performance of our algorithm as compared to the baseline is shown in Table~\ref{table:acc-cifar10} for the CIFAR-10 dataset and in Table~\ref{table:acc-mnist} for the MNIST dataset. We make a few observations. In both the cases the performance of the multiplicative weights update based algorithm is significantly better than the baseline on the minimum accuracy metric and the union attack metric. This difference is significantly higher for the MNIST dataset where both the greedy and the round robin heuristics are unstable and have much higher variances. The round robin heuristic switches among different losses much more often and pays unnecessary attention to the adversaries which it already covers well. The greedy heuristic, on the other hand, switches less frequent than our proposed algorithm. But greedy fails to address the runner-up adversary which may be almost as difficult as the chosen one, causing instability.  
Furthermore, we also notice that using the average of the last three model parameters in the multiplicative weights based method achieves slightly better performance than simply using the parameters of the last time step.
\vspace{-5mm}
\paragraph{Comparison of Training Times.} We next demonstrate the scalability of our proposed algorithm. Table~\ref{tab:train-time} shows the training time of our method, measured in wall clock time, as compared to the baselines when optimizing over all $6$ loss functions. Moreover, the first two columns represent the training times for optimizing a single loss function~($\ell_\infty$ attack) in either the pixel or the DCT basis. As can be seen our the training cost of our approach scales sublinearly with the number of representation spaces.

\vspace{-5mm}

\paragraph{On the Convexity Assumption.} Recall that the guarantees of Theorem~\ref{thm:main} apply to the algorithm in Figure~\ref{ALG:multi-robust} that requires one to produce a hypothesis that is an ensemble of the intermediate trained models. In other words, ideally one should be averaging the post-softmax outputs of the trained models. If the loss functions were convex then one could replace the ensembling with simply averaging the model weights and retain the theoretical guarantees. Even though we have non-convex losses we still make the near convexity assumption and average the model weights to produce a scalable implementation. In Table~\ref{table:convexity-cifar10} and Table~\ref{table:convexity-mnist} we compare the performance of our weight averaging strategy with that of the ideal one that ensembles the models. As can be seen the loss in making the near convexity assumption is negligible and justifies our implementation in Figure~\ref{ALG:multi-robust-scalable}.

\vspace{-5mm}
\section{Discussion}
\label{sec:conclusions}
In this work we motivated the problem of designing neural networks that are simultaneously robust to multiple types of adversarial attacks in multiple representation spaces. We provided a theoretically sound algorithm with training cost that grows sublinearly with the number of representation spaces. Our implementation is scalable and significantly outperforms strong baselines, with similar training cost to standard adversarial training.

Several future directions emerge from this work. Notice that in our proposed algorithm we use the PGD based method of~\cite{madry2017towards} to optimize the individual losses. There has been very recent work proposing faster training methods that achieve similar performance to that of the PGD method~\cite{shafahi2019adversarial, wong2020fast}. It would be interesting to incorporate them in our framework to drive down the training cost even further. The benefits of this could be significant as the number of representation spaces grows. 

While our theory applies to an ensemble of neural networks, as our experiments indicate, in practice simply averaging the model weights does as well as ensembling. This justifies our {\em near convexity} assumption. It would be interesting to study this behavior further and provide a more formal theoretical justification. Finally, we hope that future work on adversarial robustness will consider evaluating robustness of classifiers in multiple representation spaces.

\begin{table}[ht!]
\small
    \centering
    \begin{tabular}{|p{1.8cm}|p{2cm}|p{1.8cm}|}
    \hline
         &  Mult. Weights~($h=3$)~(ensemble) & Mult. Weights~($h=3$)~(wt. avg.) \\
         \hline
         Pixel ($\ell_\infty$)  & ${39.13	\pm 1.74}$ & ${40.52	\pm 1.28}$\\
         \hline
         Pixel ($\ell_2$) & $71.27	\pm 0.22$ & $71.41	\pm 0.24$\\
         \hline
         Pixel ($\ell_1$) & $46.90 \pm	1.46$ & $48.14	\pm 0.99$\\
         \hline
         DCT ($\ell_\infty$) & $42.27	\pm 2.77$ & $43.95	\pm 2.05$\\
         \hline
         DCT ($\ell_2$) & $71.24 \pm	0.12$ & $71.38	\pm 0.20$\\
         \hline
         DCT ($\ell_1$) & $42.96	\pm 1.12$ & $44.49	\pm 0.58$\\
         \hline
         Min. Accuracy & ${39.13	\pm 1.74}$ & ${40.52	\pm 1.28}$\\
         \hline
         Union Attack  & ${36.18	\pm 1.89}$ & ${37.76	\pm 1.30}$\\
         \hline
         Nat. Acc. & $80.66	\pm 0.89$ & ${80.46	\pm 0.81}$\\
         \hline
    \end{tabular}
    \caption{Comparison of the adversarial accuracies achieved on the CIFAR-10 dataset by our proposed algorithm in Figure~\ref{ALG:multi-robust-scalable} when using the average of the last three model parameters~(convexity assumption) vs. ensembling the outputs of the last three models.}
    \label{table:convexity-cifar10}
    \small
    \centering
    \begin{tabular}{|p{2cm}|p{2cm}|p{2cm}|}
    \hline
         &  Mult. Weights~($h=3$)~(ensemble) & Mult. Weights~($h=3$)~(wt. avg.)\\
         \hline
        Pixel ($\ell_\infty$) & ${66.56	\pm 13.45}$ & ${66.61	\pm 13.77}$\\
         \hline
         Pixel ($\ell_2$) & ${71.77	\pm 4.25}$ & ${71.89 \pm	3.79}$\\
         \hline
         Pixel ($\ell_1$) & ${73.66	\pm 9.47}$ & ${73.38	\pm 9.45}$\\
         \hline
         DCT ($\ell_\infty$) & ${60.32	\pm 11.34}$ & ${59.87	\pm 11.32}$\\
         \hline
         DCT ($\ell_2$) & ${84.46	\pm 4.18}$ & ${84.34	\pm 4.13}$\\
         \hline
         DCT ($\ell_1$) & ${73.13	\pm 7.33}$ & ${73.08	\pm 7.38}$\\
         \hline
         Min. Acc. & ${57.57	\pm 8.46}$ & ${57.23	\pm 8.51}$\\
         \hline
         Union Attack  & ${35.87	\pm 6.69}$ & ${35.54	\pm 6.69}$\\
         \hline
         Nat. Acc. & ${91.43	\pm 9.61}$ & ${91.44	\pm 9.62}$\\
         \hline
    \end{tabular}
    \caption{Comparison of the adversarial accuracies achieved on the MNIST dataset by our proposed algorithm in Figure~\ref{ALG:multi-robust-scalable} when using the average of the last three model parameters~(convexity assumption) vs. ensembling the outputs of the last three models.}
    \label{table:convexity-mnist}
\end{table}

\newpage

{\small
\bibliographystyle{ieee_fullname}
\bibliography{main}
}
\newpage
\input{appendix}

\end{document}

%% file: appendix.tex
\section{Appendix: Proof of Theorem~\ref{thm:main}}
\label{app:proof}
\begin{proof}
Given the weights $w_1, w_2, \dots w_k$ for the $k$ losses, we denote by $\mathbf{p}$ the normalized probabilities, i.e., $p_i = w_i/\sum_{j=1}^k w_j$. Let $\mathbf{p}_1, \mathbf{p}_2, \dots, \mathbf{p}_T$ be the sequence of probability vectors produced by the column players namely the player that chooses among the $k$ losses. Denote by $p_{t,j}$ the $j$th coordinate of the vector $\mathbf{p}_t$. Since the player is performing multiplicative weights updates, by the standard guarantee of the multiplicative weights update~(see Theorem 2 in~\cite{Kalethesis}) we get that for any $i \in [k]$ the following holds
\begin{align}
    \label{eq:mw-1}
    \frac{1}{T} \sum_{t=1}^T \sum_{j=1}^k p_{t,j} L_j(\theta_t) \geq (1-\eta) \frac{1}{T} \sum_{t=1}^T L_i(\theta_t) - 2R\frac{\log k}{\eta T}.
\end{align}
Setting $\eta = O(\epsilon/R)$ and $T = O(R^2 \frac{\log k}{\epsilon^2})$ we get that
\begin{align}
    \label{eq:mw-2}
    \frac{1}{T} \sum_{t=1}^T \sum_{j=1}^k p_{t,j} L_j(\theta_t) \geq \frac{1}{T} \sum_{t=1}^T L_i(\theta_t) - \epsilon.
\end{align}
Hence, denoting by $P$ the uniform distribution over the $T$ parameters, we get that
\begin{align}
    \label{eq:mw-3}
    \mathbb{E}_{\theta \sim P} L_i(\theta) \leq \frac{1}{T} \sum_{t=1}^T \sum_{j=1}^k p_{t,j} L_j(\theta_t) + \epsilon.
\end{align}

Next we will write down the right hand side in terms of the optimal value of the objective as in \eqref{eq:robust-obj}. Given the guarantee that we can solve the adversarial cost sensitive optimization upto additive error $\delta$, we have that for any $t \in [T]$,
\begin{align}
    \label{eq:mw-4}
    \sum_{j=1}^k p_{t,j} L_j(\theta_t) \leq \min_\theta \sum_{j=1}^k p_{t,j} L_j(\theta) + \delta.
\end{align}
Substituting into \eqref{eq:mw-3} we get
\begin{align}
    \label{eq:mw-5}
    \mathbb{E}_{\theta \sim P} L_i(\theta) &\leq \frac{1}{T} \sum_{t=1}^T \min_\theta \sum_{j=1}^k p_{t,j} L_j(\theta) + \epsilon + \delta\\
    &\leq \min_\theta \frac{1}{T} \sum_{t=1}^T  \sum_{j=1}^k p_{t,j} L_j(\theta) + \epsilon+ \delta.
\end{align}
Next for any $j \in [k]$ define $\tilde{p}_j = \frac{1}{T} \sum_{t=1}^T p_{t,j}$. Then we can rewrite the above as
\begin{align}
    \label{eq:mw-6}
    \mathbb{E}_{\theta \sim P} L_i(\theta) &\leq \frac{1}{T} \sum_{t=1}^T \min_\theta \sum_{j=1}^k p_{t,j} L_j(\theta) + \epsilon + \delta  \nonumber \\
    &\leq \min_\theta \sum_{j=1}^k \tilde{p}_{j} L_j(\theta) + \epsilon + \delta .
\end{align}
It is easy to check that $\tilde{p}_j \in [0,1]$ and $\sum_{j=1}^k \tilde{p}_j=1$. Hence we get that $\sum_{j=1}^k \tilde{p}_j L_j(\theta) \leq \min_\theta \max_j L_j(\theta)$. Substituting into \eqref{eq:mw-6} we get the guarantee of the theorem that for all $i \in [k]$,
\begin{align}
    \label{eq:mw-7}
    \mathbb{E}_{\theta \sim P} L_i(\theta) &\leq 
    \min_\theta \max_j L_j(\theta) +  \epsilon + \delta.
\end{align}

Next we prove the consequence when the loss function $L$ in \eqref{eq:robust-obj-i} is convex in its first argument. This is true for commonly used loss functions in practice such as the cross entropy loss, squared loss and the hinge loss. We denote by $f_{\theta_t}$ the network corresponding to the parameter $\theta_t$ and by $f_{\hat{\theta}}$ the ensembled network i.e., 
$$
f_{\hat{\theta}} = \frac{1}{T} \sum_{t=1}^T f_{\theta_t}.
$$
By expanding out $\mathbb{E}_{\theta \sim P} L_i(\theta)$ we get
\begin{align}
    \label{eq:mw-8}
    \mathbb{E}_{\theta \sim P} L_i(\theta) &= \frac{1}{T} \sum_{t=1}^T L_i(f_{\theta_t})\\
    &= \frac{1}{T} \sum_{t=1}^T \mathbb{E}_{(x,y) \sim D} \big[ \max_{z \in B_i(x)} L(f_{\theta_t}(\mathcal{R}^{-1}(z)),y) \big]\\
    &\geq \max_{z \in B_i(x)} \frac{1}{T} \sum_{t=1}^T \mathbb{E}_{(x,y) \sim D} \big[  L(f_{\theta_t}(\mathcal{R}^{-1}(z)),y) \big]\\
    &= \max_{z \in B_i(x)} \mathbb{E}_{(x,y) \sim D} \mathbb{E}_{\theta \sim P} L(f_{\theta}(\mathcal{R}^{-1}(z)),y)\\
    &\geq \max_{z \in B_i(x)} \mathbb{E}_{(x,y) \sim D}  L(\mathbb{E}_{\theta \sim P}f_{\theta}(\mathcal{R}^{-1}(z)),y)\\
    &= \max_{z \in B_i(x)} \mathbb{E}_{(x,y) \sim D}  L(f_{\hat{\theta}}(\mathcal{R}^{-1}(z)),y).
\end{align}
Here in the last inequality we have used the fact that $L$ is convex in its first argument. Substituting back into \eqref{eq:mw-7} we get that $f_{\hat{\theta}}$ satisfies that for all $i \in [k]$,
\begin{align}
    \label{eq:mw-9}
    L_i(f_{\hat{\theta}}) &\leq 
    \min_\theta \max_j L_j(\theta) +  \epsilon + \delta.
\end{align}
\end{proof}

\section{Appendix: Further Experiments}
\label{sec:app-expriments}
\begin{table*}[h]
\small
    \centering
    \begin{tabular}{|c|c|c|c|c|}
    \hline
         &  Greedy & Round Robin & Mult. Weights~($h=1$) & Mult. Weights~($h=3$)\\
         \hline
         Pixel ($\ell_\infty$) & $34.16	\pm 6.57$ & $28.28	\pm 1.53$ & ${34.01	\pm 6.38}$ & $\mathbf{37.24	\pm 1.89}$\\
         \hline
         Pixel ($\ell_2$) & $64.74	\pm 1.73$ & $\mathbf{67.33	\pm 0.27}$ & $64.97	\pm 3.03$ & $\mathbf{66.94	\pm 0.28}$\\
         \hline
         Pixel ($\ell_1$) & $45.26	\pm 3.11$ & $\mathbf{53.08	\pm 0.72}$ & $43.46 \pm	4.99$ & $46.25	\pm 1.44$\\
         \hline
         DCT ($\ell_\infty$) & ${42.64	\pm 9.02}$ & $\mathbf{43.62	\pm 1.74}$ & $37.17	\pm 6.92$ & $39.93	\pm 3.01$\\
         \hline
         DCT ($\ell_2$) & $64.66	\pm 1.71$ & $\mathbf{67.26	\pm 0.38}$ & $65.03 \pm	2.88$ & $66.98	\pm 0.30$\\
         \hline
         DCT ($\ell_1$) & $37.07	\pm 6.84$ & $\mathbf{41.01	\pm 0.81}$ & $39.16	\pm 5.86$ & $\mathbf{42.52	\pm 1.17}$\\
         \hline
         Min. Accuracy & $32.50	\pm 5.31$ & $28.28	\pm 1.53$ & $\mathbf{34.01	\pm 6.38}$ & $\mathbf{37.24	\pm 1.89}$\\
         \hline
         Union Attack  & $30.29	\pm 4.24$ & $28.07	\pm 1.48$ & $\mathbf{31.61	\pm 6.07}$ & $\mathbf{34.63	\pm 2.02}$\\
         \hline
         Nat. Acc. & $78.56	\pm 1.46$ & $\mathbf{81.80	\pm 0.38}$ & $79.64	\pm 0.51$ & $\mathbf{80.66	\pm 0.89}$\\
         \hline
    \end{tabular}
    \caption{Comparison of the adversarial accuracies achieved on the CIFAR-10 dataset by the greedy algorithm, the round robin algorithm and our proposed algorithm in Figure~\ref{ALG:multi-robust-scalable}.}
    \label{table:acc-cifar10-20-restarts}
\end{table*}
\begin{table*}[h]
\small
    \centering
    \begin{tabular}{|c|c|c|c|c|}
    \hline
         &  Greedy & Round Robin & Mult. Weights~($h=1$) & Mult. Weights~($h=3$)\\
         \hline
         Pixel ($\ell_\infty$) & $53.50	\pm 24.05$ & $59.60	\pm 24.55$ & $\mathbf{60.88	\pm 17}$ & $\mathbf{60.01	\pm 16.61}$\\
         \hline
         Pixel ($\ell_2$) & $7.50	\pm 4.48$ & $6.04	\pm 4.27$ & $\mathbf{63.32	\pm 8.25}$ & $\mathbf{63.47 \pm	7.72}$\\
         \hline
         Pixel ($\ell_1$) & $63.60	\pm 30.36$ & $41.98	\pm 22.87$ & $\mathbf{72.23	\pm 10.54}$ & $\mathbf{72.51	\pm 9.68}$\\
         \hline
         DCT ($\ell_\infty$) & $53.80	\pm 24.53$ & $53.24	\pm 22.41$ & $\mathbf{55.29	\pm 12.16}$ & $\mathbf{54.81	\pm 13.09}$\\
         \hline
         DCT ($\ell_2$) & $9.18	\pm 5.19$ & $5.90	\pm 5.71$ & $\mathbf{79.97	\pm 5.92}$ & $\mathbf{79.50	\pm 6.50}$\\
         \hline
         DCT ($\ell_1$) & $65.45	\pm 31.95$ & $36.81	\pm 21.10$ & $\mathbf{71.74	\pm 8.49}$ & $\mathbf{72.51	\pm 7.33}$\\
         \hline
         Min. Accuracy & $7.31	\pm 4.38$ & $5.19	\pm 4.72$ & $\mathbf{52.53	\pm 11.31}$ & $\mathbf{53.29	\pm 12.73}$\\
         \hline
         Union Attack  & $3.37	\pm 3.10$ & $1.97	\pm 3.03$ & $\mathbf{27.65	\pm 7.06}$ & $\mathbf{28.09	\pm 9.38}$\\
         \hline
         Nat. Acc. & $77.16	\pm 37.64$ & $60.93	\pm 32.31$ & $\mathbf{91.00	\pm 10.50}$ & $\mathbf{91.43	\pm 9.61}$\\
         \hline
    \end{tabular}
    \caption{Comparison of the adversarial accuracies achieved on the MNIST dataset by the greedy algorithm, the round robin algorithm and our proposed algorithm in Figure~\ref{ALG:multi-robust-scalable}.}
    \label{table:acc-mnist-20-restarts}
\end{table*}

In this section we present the results of evaluating our trained multiplicative weights method based algorithm from Figure~\ref{ALG:multi-robust-scalable}, as well as the greedy and the round robin heuristics, against the PGD based attack where the PGD method is run with $20$ random restarts in order to find an adversarial example. Tables~\ref{table:acc-cifar10-20-restarts} shows the results for the CIFAR-10 dataset and Table~\ref{table:acc-mnist-20-restarts} shows the results for the MNIST dataset. Similar to the results presented in Section~\ref{sec:experiments}, the multiplicative weights method significantly outperforms the baselines on both the minimum accuracy metric and accuracy against a union attack.